%% file: arxiv.tex
\documentclass[11pt]{article}
\usepackage{enumerate}
\usepackage{fullpage}
\usepackage{algorithm}
\usepackage[noend]{algorithmic}

\usepackage{xspace}
\usepackage[disable]{todonotes}

\newcommand{\todoa}[2][]{\todo[linecolor=green,backgroundcolor=green!25,bordercolor=green,#1]{A: #2}\xspace}

\usepackage{amsmath,amsthm,amsfonts,amssymb}
\usepackage{amsmath}
\usepackage{hyperref}
\usepackage{thm-restate}
\usepackage{color}
\usepackage{mathrsfs}
\usepackage{enumitem}
\usepackage{bm}
\usepackage{multirow}
\usepackage{booktabs}
\usepackage{makecell}
\usepackage{graphicx}
\usepackage{subfigure}
\usepackage{caption}
\usepackage{thmtools}
\usepackage{thm-restate}
\usepackage{setspace}
\usepackage{makecell}
\definecolor{light-gray}{gray}{0.85}
\usepackage{colortbl}
\usepackage{xcolor}
\usepackage{hhline}
\usepackage{xspace} 
\usepackage{mathtools} 
\usepackage{cancel}
\usepackage{dsfont}
\usepackage[numbers]{natbib}

\include{packages}
\usepackage{amsfonts}

\include{notations}
\usepackage{times}

\begin{document}

 \title{\fontsize{16pt}{16pt}
 \textbf{Optimistic Natural Policy Gradient: a Simple Efficient Policy Optimization Framework for Online RL
}
 }
 
 \date{}
 
 \author{ 
  Qinghua Liu\footnote{This work was done during QL's internship at DeepMind.} \\
  Princeton University \\
  \texttt{qinghual@princeton.edu} \\
  \and
  Gell\'ert Weisz \\
  Google DeepMind \\
  \texttt{gellert@deepmind.com} \\
  \and
    Andr\'as Gy\" orgy\\
  Google DeepMind \\
  \texttt{agyorgy@deepmind.com} \\
  \and
   Chi Jin \\
  Princeton University \\
  \texttt{chij@princeton.edu} \\
   \and
   Csaba Szepesv{\'a}ri \\
  Google DeepMind and University of Alberta \\
  \texttt{szepesva@ualberta.ca} \\
}

 \maketitle

\input{Sections/abs}

\input{Sections/intro}

\input{Sections/prelim}

 \input{Sections/opt_npg}

\input{Sections/examples}

 \input{Sections/conc}

\bibliographystyle{plainnat}
\bibliography{ref}

\newpage
\appendix

 \input{Sections/proofs}

\input{Sections/app}

\end{document}

%% file: notations.tex
\usepackage{mathtools}
\usepackage{microtype}
\usepackage{graphicx}
\usepackage{subfigure}
\usepackage{booktabs} 

\usepackage{hyperref}

\usepackage{natbib}
\usepackage{color}
\usepackage{amsmath}
\usepackage{bm}
\usepackage{amsthm}
\usepackage{amsfonts}

\newcommand{\piout}{\pi^{\rm out}}
\renewcommand{\P}{\mathbb{P}}
\newcommand{\E}{\mathbb{E}}
\newcommand{\I}{\mathbf{I}}
\newcommand{\R}{\mathbb{R}}

\newcommand{\cD}{\mathcal{D}}
\newcommand{\cO}{\mathcal{O}}
\newcommand{\cT}{\mathcal{T}}
\newcommand{\cS}{\mathcal{S}}
\newcommand{\cA}{\mathcal{A}}
\newcommand{\cF}{\mathcal{F}}
\newcommand{\cX}{\mathcal{X}}

\newcommand{\cB}{\mathcal{B}}
\newcommand{\cL}{\mathcal{L}}
\newcommand{\cG}{\mathcal{G}}

\renewcommand{\th}{^{\rm th}}
\newcommand{\argmax}{{\rm argmax}}
\newcommand{\argmin}{{\rm argmin}}

\newcommand{\poly}{{\mathrm{poly}}}

\newcommand{\trans}{^{\rm T}}

\newcommand{\setto}{\leftarrow}
\newcommand{\oV}{\overline{V}}
\newcommand{\oQ}{\overline{Q}}
\newcommand{\otheta}{\overline{\theta}}

\newtheorem{definition}{Definition}
\newtheorem{condition}{Condition}
\newtheorem{theorem}{Theorem}
\newtheorem{assumption}{Assumption}
\newtheorem{observation}{Observation}

\newtheorem{lemma}{Lemma}
\newtheorem{proposition}{Proposition}

\newcommand{\oppo}{\textsc{Optimistic NPG}}
\newcommand{\ope}{\textsc{OPE}}

\newcounter{subroutine}
\makeatletter
\newenvironment{subroutine}[1][htb]{%
  \let\c@algorithm\c@subroutine
  \renewcommand{\ALG@name}{Subroutine}
  \begin{algorithm}[#1]%
  }{\end{algorithm}
}
\makeatother

%% file: Sections/abs.tex
\begin{abstract}

While policy optimization algorithms have played an important role in recent empirical success of Reinforcement Learning (RL), the existing theoretical understanding of policy optimization remains rather limited---they are either restricted to tabular MDPs or suffer from highly suboptimal sample complexity, especially in online RL where exploration is necessary.
This paper proposes a simple efficient policy optimization framework---\oppo~for online RL. \oppo~can be viewed as a simple combination of the classic natural policy gradient (NPG) algorithm \citep{kakade2001natural} and an  optimistic policy evaluation subroutine to encourage exploration. For $d$-dimensional linear MDPs, \oppo~is computationally efficient, and learns an $\epsilon$-optimal policy within $\tilde{\cO}(d^2/\epsilon^3)$ samples, which is the first computationally efficient algorithm whose sample complexity has the optimal  
dimension dependence $\tilde{\Theta}(d^2)$. It also improves over state-of-the-art results of policy optimization algorithms \citep{zanette2021cautiously} by a factor of $d$. For general function approximation that subsumes linear MDPs, \oppo, to our best knowledge, is also the first policy optimization algorithm that achieves the polynomial sample complexity for learning near-optimal policies.


\end{abstract}

%% file: Sections/intro.tex

\section{Introduction}

Policy optimization algorithms \citep{schulman2017proximal,schulman2015trust} with neural network function approximation have played an important role in recent empirical success of  reinforcement learning (RL), such as robotics \citep{finn2016guided}, games \citep{berner2019dota} and large language models \citep{chatgpt}.  Motivated by the empirical success, the theory community made a large effort to 
design provably efficient policy optimization algorithms that work in the presence of linear function approximation \citep{agarwal2021theory,bhandari2019global,liu2019neural,neu2017unified,abbasi2019politex,agarwal2020pc,zanette2021cautiously,shani2020optimistic,cai2020provably}.
Early works focused on proving that  policy optimization algorithms are capable to learn near-optimal policies using a polynomial number of samples under certain reachability (coverage) assumptions.
 \citep[e.g.,][]{agarwal2021theory,bhandari2019global,liu2019neural,neu2017unified,abbasi2019politex}.
While this was good for laying down the foundations for future work,  
the reachability assumptions  basically imply that the state space is already well-explored or rather easy to explore, which avoids the challenge of performing strategic exploration --- one of the central problems in both empirical and theoretical RL. 

To address this limitation,  later works \citep{agarwal2020pc,zanette2021cautiously}  proposed  policy optimization algorithms that enjoy polynomial sample-complexity guarantee without making any reachability assumption, but at the cost of either  complicating the algorithm design (and analysis) with  various  tricks or getting highly suboptimal sample-complexity guarantees. 
 For example, the PC-PG algorithm \citep{agarwal2020pc}, which, to our knowledge was the first policy optimization algorithm for learning linear MDPs without reachability assumptions, 
 requires  $\tilde{\cO}(\poly(d)/\epsilon^{11})$ samples to learn an $\epsilon$-optimal policy for $d$-dimensional linear MDPs. 
That $\tilde{\cO}(1/\epsilon^{11})$ samples were necessary for this task 
 is highly unlikely.
 Indeed,  \citet{zanette2021cautiously} greatly improved this sample complexity to $\tilde{\cO}(d^3/\epsilon^3)$ at the cost of considerably  complicating the algorithm design and the analysis. Moreover, we are not aware of efficient guarantees of policy optimization for generic function approximation, which is rich enough to subsume linear MDPs.
This motivates us to ask the following question:
 \begin{center}
    \textsl{Can we design a \textbf{simple, general} policy optimization algorithm \\ with \textbf{sharp} sample-complexity guarantees 
    in the \textbf{exploration} setting\footnote{By ``exploration setting``, we mean a setup where there is no simulator nor reachability assumption and a learner can only influence the evolution of environmental states by taking actions.}?}\todoa{The "exploration setting" should be defined}
 \end{center}
In particular, we aim to achieve sharper sample complexity than \citet{zanette2021cautiously} in the linear MDP setting, and achieve low-degree polynomial sample complexity in the general function approximation setting.
This paper answers the highlighted question affirmatively by making the following contributions:


\begin{table*}[!t]
    \renewcommand{\arraystretch}{1.6} 
    \centering
    \caption{\label{table:summary_results} A comparison of sample-complexity results for linear MDPs.}
    \begin{tabular}{|c|c|c|c|}
      \hline
      \textbf{Algorithms} & \textbf{Sample Complexity} & \textbf{\thead{Computationally\\Efficient}} & \textbf{\thead{Policy\\Optimization}} \\ \hline\hline
      \thead{\citet{zanette2020learning}\\\citet{jin2021bellman}} & $\tilde{\Theta}(d^2/\epsilon^2)$   &  $\times$ & $\times$\\\hline
      \thead{\citet{he2022nearly}\\\citet{agarwal2022vo}\\ \citet{wagenmaker2022reward}} & $\tilde{\cO}(d^2/\epsilon^2+d^{\ge 4}/\epsilon)$   &  $\checkmark$ & $\times$\\\hline
      \cite{agarwal2020pc} & $\tilde{\cO}(\poly(d)/\epsilon^{11})$ & $\checkmark$ & $\checkmark$ \\ \hline
      \citet{zanette2021cautiously}  &  $\tilde{\cO}(d^3/\epsilon^3)$  & $\checkmark$ & $\checkmark$ \\ \hline
      {\cellcolor{light-gray} Optimistic NPG (\textbf{this work})} &  {\cellcolor{light-gray} $\tilde{\cO}(d^2/\epsilon^3)$}  & {\cellcolor{light-gray} $\checkmark$} & {\cellcolor{light-gray}$\checkmark$} \\\hline
    \end{tabular}
\end{table*}

\begin{itemize}
    \item \textbf{Sharper rate.} We propose a computationally efficient policy optimization algorithm---\oppo~with sample complexity  
    $$\tilde{\cO}(d^2/\epsilon^3)$$
    for learning an $\epsilon$-optimal policy in 
    an online fashion while interacting with a 
    $d$-dimensional
    linear MDP.
    This result improves over the best previous one \citep{zanette2021cautiously} in  policy optimization by a factor of $d$. 
    Moreover, to our knowledge, this is the first computationally efficient algorithm to achieve the optimal quadratic dimension dependence. 
    Before moving on, we remark that previous FQI-style algorithms \citep{zanette2020learning,jin2021bellman} achieve the optimal 
     sample complexity $\tilde{\Theta}(d^2/\epsilon^2)$ \citep{zanette2020learning}
      but they are   computationally \emph{inefficient} due to the mechanism of global optimism  in  
    their   algorithm design. 
    Several very recent works
    \citep{he2022nearly,agarwal2022vo,wagenmaker2022reward} achieve the sample complexity 
    $ \tilde{\cO}(d^2/\epsilon^2+d^{\ge 4}/\epsilon) $ using computationally efficient algorithms. 
    Compared to our work, the aforementioned rate has worse dependence on $d$ and better dependence on $\epsilon$.
    Nonetheless, our result is better in the practically important regime where the feature dimension $d$ is typically very large and the target accuracy $\epsilon$ is not too small. 
    To summarize, the sample complexity of \oppo~strictly improves over the best existing policy optimization algorithms and is not dominated by that of any existing computationally efficient algorithm. 
    
    To our best knowledge, this paper also achieves the first polynomial sample-complexity guarantee for policy optimization under general function approximation.\todoa{General?}

    
     \item \textbf{Simple on-policy algorithm.} 
     At a high level, \oppo~is almost an \emph{on-policy} version of natural policy gradient (NPG) \citep{kakade2001natural} 
     with Boltzmann policies that use linear function approximation 
     for optimistic action-value estimates. 
     By ``on-policy'', we mean the transition-reward data used to improve the current policy are obtained exactly by executing the current policy.
	However, the algorithm has a tuneable parameter (denoted by $m$ in Algorithm \ref{alg:oppo}) that allows it to reuse data sampled from 
	earlier policies. 
	By using such data, the algorithm becomes \emph{off-policy}.
	The optimal choice of the tuneable parameter is dictated by theoretical arguments.
	Interestingly, our analysis shows that 
	even the purely on-policy version of the algorithm (i.e., set $m=1$ in Algorithm \ref{alg:oppo}) enjoys
	a well-controlled sample complexity of $\tilde{\cO}(d^2/\epsilon^4)$.
	To the best of our knowledge, this is the first time an on-policy 
	method is shown to have polynomial sample complexity in the exploration setting, given that all previous policy optimization or Q-learning style or model-based algorithms \citep[e.g.,][etc]{jin2020provably,zanette2020learning,zanette2021cautiously,agarwal2020pc,cai2020provably,shani2020optimistic}  are off-policy.  
 \item \textbf{New proof techniques.} To achieve the improved rate, we introduce several  new ideas in the proof, including but not limited to  (a) exploiting the softmax
parameterization of the policies to reuse data and improve sample efficiency (Lemma 3), (b) controlling \emph{on-policy} uncertainty  (Lemma 4) instead of cumulative uncertainty as in previous off-policy works, and  (c) using a bonus term that is smaller than those in previous works by a factor of $\sqrt{d}$ (see Lemma 2 and the discussions that follow).

\end{itemize}

\subsection{Related works}
 
Since this paper studies policy optimization algorithms in the setting of linear MDPs, below we  restrict our focus to previous theoretical works on either  policy optimization or linear MDPs.

 \paragraph{Policy optimization.}
This work is inspired by and  builds upon two recent works \citep{shani2020optimistic,cai2020provably} that combine NPG with bonus-based optimism to handle exploration. 
In terms of algorithmic design, this paper (\oppo) utilizes on-policy fresh data for value estimation while \citet{shani2020optimistic,cai2020provably} are off-policy and reuse \emph{all} the historical data.
In terms of theoretical guarantees, \citet{shani2020optimistic} and \citet{cai2020provably} only study tabular MDPs and  linear mixture MDPs \citep{zhou2021nearly} respectively, while this paper considers the more challenging setting of linear MDPs \citep{jin2020provably} (in our view, linear MDPs are more challenging as there the number of model parameters scale with the number of states).
 We remark that due to some subtle technical challenge, so far it still remains unknown whether it is possible to generalize the analysis of \citet{cai2020provably} to  handle linear MDPs. 
\citet{agarwal2020pc,zanette2021cautiously} derive the first line of policy optimization results for RL in linear MDPs without any reachability- (or coverage-) style assumption.\todoa{Added hyphens -- alternatively, remove hyphens and "style"} 
Compared to \citet{agarwal2020pc,zanette2021cautiously}, our work\todoa{"This paper" is used somewhat ambigously throught} considers the same setting but designs a simpler algorithm with cleaner analysis and sharper sample-complexity guarantee. Nevertheless, we remark that in certain model-misspecification settings, the algorithms of \citet{agarwal2020pc,zanette2021cautiously} can potentially achieve stronger guarantees. Since we focus on the well-specified setting (that is, the environment is perfectly modeled by a linear MDP), we refer interested readers to \citet{agarwal2020pc} for more details.\todoa{This part is not entirely clear} 
Finally, there have been a long line of works \citep[e.g.,][etc]{agarwal2021theory,bhandari2019global,liu2019neural,neu2017unified,abbasi2019politex} that study policy optimization under reachability (or coverage) assumptions, which eliminates the need for performing strategic exploration. 
Throughout this paper we do not make any such assumption  and  directly tackle the exploration challenge.
 
 \paragraph{Linear MDPs.}
 
For linear MDPs, \citet{jin2020provably} proposed a computationally efficient algorithm (LSVI-UCB) with  
$\tilde{\cO}(d^3/\epsilon^2)$ sample complexity. 
Later on, \citet{zanette2020learning} utilized the idea of global optimism to obtain the optimal  sample complexity $\tilde{\Theta}(d^2/\epsilon^2)$  at the cost of sacrificing  computational efficiency.
Recently, \citet{he2022nearly,agarwal2022vo,wagenmaker2022reward} designed new computationally efficient algorithms that can achieve  $\tilde{\cO}(d^2/\epsilon^2+d^{\ge 4}/\epsilon)$ sample complexity. 
Compared to the above works, the sample complexity of \oppo~is not strictly worse than that of any known computationally efficient algorithm. In fact, it is the only computationally efficient method to achieve the optimal dimension dependence.
Nonetheless, for learning a near-optimal policy with a  vanishing suboptimality $\epsilon$, the current sample complexity of \oppo~is loose by a factor of $\epsilon^{-1}$.

%% file: Sections/prelim.tex

\section{Preliminaries}

\paragraph{MDP.}
We consider the model of episodic Markov Decision Processes (MDPs).
Formally, an MDP  is defined by a tuple $(\cS,\cA,\P,R,H)$ where $\cS$ denotes the set of states, $\cA$ denotes the set of actions, both of which are assumed to be finite,\todoa{Added finiteness assumptions.} $\P=\{\P_h\}_{h\in[H]}$ denotes the set of transition probability functions so that $\P_h(s'\mid s,a)$ is equal to the probability of transitioning to state $s'$ given that action $a$ is taken at state $s$ and step $h$, $R=\{R_h\}_{h\in[H]}$ denotes the collection of expected reward functions so that $R_h(s,a)$ is equal to the expected reward to be received if action $a$ is taken at state $s$ and step $h$, and $H$ denotes the length of each episode.  
An agent interacts an MDP in the form of episodes. Formally, we assume without loss of generality that each episode always starts from a \emph{fixed} initial state  $s_1$. 
At the $h\th$  step of this episode, the agent first observes the current state $s_h$, then takes action $a_h$ and receives reward $r_h(s_h,a_h)$ satisfying 
\[
\E[r_h(s_h,a_h) \mid s_h,a_h]=R_h(s_h,a_h)\,.
\] After that, the environment transitions to 
\[
s_{h+1}\sim \P_h(\cdot\mid s_h,a_h)\,. 
\]
The current episode terminates immediately once $r_H$ is received. 
Throughout this paper, we assume $\P$ and $R$ are unknown to the learner.

\paragraph{Linear MDP.}
A $d$-dimensional linear MDP \citep{jin2020provably} is defined by  two sets of feature mappings, $\{\phi_h\}_{h\in[H]}\subseteq(\cS\times\cA\rightarrow \R^d)$ and $\{\psi_h\}_{h\in[H]}\subseteq(\cS\rightarrow \R^d)$, and a set of vectors $\{w^\star_h\}_{h\in[H]}\subseteq\R^d$ so that the transition probability functions can be represented as bilinear functions of feature mappings $\{\phi_h\}_{h\in[H]}$ and $\{\psi_h\}_{h\in[H]}$, and the reward functions can be represented as linear functions of $\{\phi_h\}_{h\in[H]}$. Formally, we have  that for all $(s,a,s')\in\cS\times\cA\times\cS$:
\begin{align*}
\P_h(s'\mid s,a) 
& = \langle \phi_h(s,a), \psi_h(s')\rangle,\\ 
R_h(s,a) &=\langle \phi_h(s,a), w^\star_h\rangle.
\end{align*}
For the purpose the regularity, linear MDPs also require that 
$$
\max_{h,s,a}\|\phi_h(s,a)\|_2 \le 1,  \quad \max_h \|w^\star_h\|_2 \le \sqrt{d},
$$
and for any function $V_{h+1}:\cS\rightarrow[0,1]$, 
$$
\left\|\int_{s\in\cS}V_{h+1}(s) \psi_{h+1}(s) ds \right\|_2 \le \sqrt{d}.
$$
Throughout this paper, we assume only $\{\phi_h\}_{h\in[H]}$ is available to the learner while 
$\{\psi_h\}_{h\in[H]}$ and 
$\{w^\star_h\}_{h\in[H]}$ are not.

\paragraph{Policy and value.} 

A (Markov) policy is a set of conditional probability functions $\pi=\{\pi_h\}_{h\in[H]}$ so that  $\pi_h(\cdot\mid s)\in\Delta_\cA$ gives the distribution over the action set conditioned on the current state $s$ at step $h$.
We define the V-value functions of policy $\pi$ by $\{V^\pi_h\}_{h\in[H]}\subseteq (\cS\rightarrow \R)$ so that $V_h^\pi(s)$ is equal to the expected cumulative reward an agent will receive if she follows policy $\pi$ starting from state $s$ and step $h$. Formally,\todoa{Add $s'_{h+1} \sim \P_h(\cdot\mid s'_h,a'_h)$?} 
$$
V^\pi_h(s)= \E\left[ \sum_{h'=h}^H r_{h'}(s_{h'},a_{h'})\mid s_h=s, a_{h'}\sim \pi_{h'}(s_{h'}), s'_{h+1} \sim \P_h(\cdot\mid s'_h,a'_h) \right],
$$
where the expectation is with respect to the randomness  of the transition, the reward and the policy. Similarly, we can define the Q-value functions of policy $\pi$ by $\{Q^\pi_h\}_{h\in[H]}\subseteq (\cS\times\cA\rightarrow \R)$ so that $Q_h^\pi(s,a)$ is equal to the expected cumulative reward an agent will receive if she follows policy $\pi$ starting from taking action $a$ at state $s$ and step $h$. Formally, 
\begin{align*}
Q^\pi_h(s,a)
 = \E\Biggl[ \sum_{h'=h}^H r_{h'}(s_{h'},a_{h'})\mid  
(s_h,a_h)=(s,a), a_{h'}\sim \pi_{h'}(s_{h'}),s'_{h+1} \sim \P_h(\cdot\mid s'_h,a'_h) \Biggr].
\end{align*}
We denote by $\pi^\star=\{\pi^\star_h\}_{h\in[H]}$ the optimal policy such that $\pi^\star  \in \argmax_\pi V^\pi_h(s)$ for all $(s,h)\in\cS\times[H]$.
By backward induction, one can easily prove that there always exists an optimal Markov policy. For simplicity of notations, we denote by $V^\star_h=V^{\pi^\star}_h$ and $Q^\star_h=Q^{\pi^\star}_h$ the optimal value functions. Note that given an MDP, the optimal value functions are unique despite that there may exist multiple optimal policies.\todoa{Reference? QL: I feel this is a folklore result for MDPs. Not sure which paper to cite.}

\paragraph{Learning objective.} 
The objective of this paper is to design an efficient policy optimization algorithm to  learn an $\epsilon$-optimal policy $\pi$ such that $V^\pi_1(s_1)\ge V^\star_1(s_1)-\epsilon$.
Here the optimality is only  measured by the value at the initial state $s_1$,  because (a) each episode always starts from $s_1$, and (b) this paper  studies the online exploration setting without access to a simulator, which means some states might be unreachable and learning optimal policies starting from those states is in general impossible. 

%% file: Sections/opt_npg.tex
\section{Optimistic Natural Policy Gradient}
In this section, we present the algorithm \oppo~(Optimistic Natural Policy Gradient) and its theoretical guarantees.

\subsection{Algorithm} The pseudocode of \oppo~is provided in Algorithm \ref{alg:oppo}.
At a high level, the algorithm consists of the following three key modules.
\begin{itemize}
    \item \textbf{Periodic on-policy data collection} (Lines~\ref{line:sample-0}-\ref{line:sample-1}): Similarly to the empirical PPO algorithm \citep{schulman2017proximal} ,\todoa{Reference?}  \oppo~discards all the old data, and
     executes the \emph{current} policy $\pi^k$ to collect a batch of fresh data $\cD^k$ after every $m$ steps of policy update. These  data will later be  used to evaluate and improve the policies in the next $m$ steps.
    Noticeably, this \emph{on-policy} data mechanism is very different from most existing works in theoretical RL, where they either need to keep the historical data or have to rerun historical policies to refresh the dataset for the technical purpose of elliptical potential arguments in proofs.
    In comparison, \oppo~only uses  fresh data collected by the current (or very recent) policy, which resembles practical policy optimization algorithms such as PPO \citep{schulman2017proximal} and TRPO \citep{schulman2015trust}.\todoa{References?}

\item \textbf{Optimistic policy evaluation}  (Line~\ref{line:pe}): Given the above collected dataset $\cD^k$, we estimate the value functions of the current policy $\pi^k$ by invoking Subroutine \ope~(optimistic policy evaluation). 
In Section \ref{sec:examples}, we show how to implement Subroutine \ope~for tabular MDPs, linear MDPs and RL problems of low eluder dimension.


    \item \textbf{Policy update} (Line~\ref{line:pu}): Given the optimistic value estimates $\{\oQ^k_h\}_{h\in[H]}$ of $\pi^k$, \oppo~performs one-step mirror ascent from $\pi_h^k(\cdot \mid s)$ with gradient $\oQ^k_h(s,\cdot)$  to obtain
    $\pi_h^{k+1}(\cdot \mid s)$ at each $(h,s)$. 
    Importantly, this step can be implemented in a computationally efficient way, because by the update rule of mirror ascent $\pi^{k}_h(\cdot\mid s)\propto \exp(\eta\sum_{t=1}^{k-1} \oQ_h^t(s,\cdot))$, we only need to store $\{\{\oQ_h^t\}_{h\in[H]}\}_{t\in[K]}$, from which any $\pi^{k}_h(\cdot\mid s)$ can be computed on the fly.\todoa{Don't we only need to store the sums $\sum_{t=1}^{k-1} \oQ_h^t(s,a)$ for all $a$? \\
    QL: To store the average of $k$ Q-functions, there seems no better way than storing them separately?  }
\end{itemize}

\begin{algorithm}[t]
\caption{\oppo}
\label{alg:oppo}
\begin{algorithmic}[1]
\STATE \textbf{input}:
number of iterations $K$, period of collecting fresh data  $m$, batch size $N$, learning rate $\eta$
\STATE \textbf{initialize:} for all $(h,s)\in[H]\times\cS$ set $\pi^1_h(\cdot\mid s)={\rm Uniform}(\mathcal{A})$
\FOR{$k=1,\ldots,K$}
\IF{$k\equiv 1 (\mbox{mod } m)$} \label{line:sample-0}
\STATE
$\cD^k\setto$ $\{N$ fresh trajectories $\stackrel{\text{i.i.d.}}{\sim}\pi^k\}$
\ELSE
\STATE $\cD^k  \setto \cD^{k-1}$
\ENDIF \label{line:sample-1}
\STATE
$\{\oQ^k_h\}_{h\in[H]} \leftarrow\ope(\pi^k, \cD^k)$
\label{line:pe}
\STATE for all $(h,s)\in[H]\times\cS$ update  $\pi^{k+1}_h(\cdot\mid s) \propto \pi^{k}_h(\cdot\mid s) \cdot \exp( \eta \cdot \oQ_h^k(s,\cdot))$
\label{line:pu}
\ENDFOR
\STATE \textbf{output}: $\piout$ that is sampled uniformly at random from $\{\pi^k\}_{k\in[K]}$
\end{algorithmic}
\end{algorithm}

\subsection{Theoretical guarantee}

\oppo~is a generic policy optimization meta-algorithm, and we can show it provably learns near-optimal policies as long as subroutine \ope~satisfies the following condition.

\begin{condition}[Requirements for \ope]
\label{cond:ope}
Suppose parameter $m$ and $\eta$ in \oppo~satisfy $m\le(\eta H^2)^{-1}$. 
Then with probability at least $1-\delta$, $\ope(\pi^k, \cD^k)$ returns $\{\oQ^k_h\}_{h\in[H]}\subseteq(\cS\times\cA\rightarrow[0,H])$ that satisfy the following properties for all $k\in[K]$.
\begin{enumerate}[label=(1\Alph*), topsep=0pt, itemsep=0pt]
    \item \textbf{\emph{Optimism:}} For all  $(s,a,h)\in\cS\times\cA\times[H]$ 
    $$\oQ_h^k(s,a) \ge (\cT^{\pi^k}_h \oQ_{h+1}^k)(s,a)$$
    where  $(\cT^{\pi}_h f)(s,a)
    = R_h(s,a)
 +\E\left[ f(s',a')\mid s'\sim\P_h(\cdot \mid s,a),a'\sim\pi_{h+1}(s')\right].$\label{cond:opt}
    \item \textbf{\emph{Consistency:}}
    There exists an absolute complexity measure $L\in\R^+$ such that 
    $$
    \oV_1^k(s_1)-V_1^{\pi^k}(s_1) \le \sqrt{(L/N)\times\log^2(NKL/\delta)},$$
    where $\oV_1^k(s_1)=\E\left[\oQ_1^k(s_1,a_1)\mid a_1\sim \pi^k(s_1)\right].$
    \label{cond:consist}
\end{enumerate}
\end{condition}
Condition \ref{cond:opt} requires that the Q-value estimate returned by \ope~satisfies the Bellman equation under policy $\pi^k$ optimistically. Requiring optimism is very common in analyzing RL algorithms, as most algorithms rely on the principle of optimism in face of uncertainty to handle exploration. 
Condition \ref{cond:consist} requires that the degree of over-optimism at initial state $s_1$ is roughly of order $\sqrt{1/N}$ with respect to the number of samples $N$. This is intuitive since as more data are collected from policy $\pi^k$ or some policy similar to $\pi^k$ (as is enforced by the precondition $m\le(\eta H^2)^{-1}$), the value estimate at the initial state should be more accurate. In Section \ref{sec:examples}, we will provide concrete instantiations of Subroutine \ope~for tabular MDPs, linear MDPs and RL problems of low eluder dimension, all of which satisfy Condition \ref{cond:ope} with mild complexity measure $L$.

Under Condition \ref{cond:ope}, we have the following theoretical guarantees for \oppo, which we prove in Appendix~\ref{sec:proof-sketch}.

\begin{theorem}\label{thm:main}
Suppose Condition \ref{cond:ope} holds.
In Algorithm \ref{alg:oppo}, if we choose
\begin{align*}
&K =\Theta\left(\frac{H^4\log|\cA|}{\epsilon^2}\right),\quad
N=\Theta\left(\frac{L \log^2(LK/\delta)}{\epsilon^2}\right), 
&\eta =\Theta\left(\frac{\epsilon}{H^3}\right),\quad
\quad m\le \Theta\left(\frac{H}{\epsilon}\right),
\end{align*}
 then with probability at least $1/2$, $\piout$  is $\cO(\epsilon)$-optimal. 
\end{theorem}
Below we emphasize two special choices of $m$ (the period of sampling fresh data) in Theorem \ref{thm:main}:

\begin{itemize}
    \item When choosing $m=1$, \oppo~is purely \emph{on-policy} as it only uses data sampled from the current policy to perform policy optimization. In this case, the total sample complexity is
    $$
    \frac{\text{\# iteration}}{\text{period of sampling}}\times\text{batch size}=
    \frac{K}{m}\times N= \tilde{\Theta}\left( \frac{L H^4}{\epsilon^4}\right),
    $$
    which, to our knowledge, is the first polynomial sample complexity guarantee for purely on-policy algorithms.
    \item When choosing $m=[H/\epsilon]$, we obtain sample complexity
     $$
     \frac{\text{\# iteration}}{\text{period of sampling}}\times\text{batch size}
     = \frac{K}{m}\times N
     = \tilde{\Theta}\left( \frac{L H^3}{\epsilon^3}\right),
    $$
    which improves  over the above purely on-policy version by a factor of $H/\epsilon$. 
    In Section \ref{subsec:linear}, we will specialize this result to linear MDPs to derive a sample complexity that improves the 
    best existing one in policy optimization  \citep{zanette2021cautiously} by a factor of $d$.
\end{itemize}
Finally we remark that for cleaner presentation, we state Theorem \ref{thm:main} for a fixed, constant failure probability.
However, the result is easy to extend to the case when the failure probability is some arbitrary value of $\delta\in (0,1)$ at the  expense of increasing the sample complexity by a factor of $\log(1/\delta)$: one simply need to run Algorithm \ref{alg:oppo} for $\log(1/\delta)$ times, estimate the values of every output policy to accuracy $\epsilon$, and then pick the one with the highest estimate.

%% file: Sections/examples.tex
\section{Examples}
\label{sec:examples}

In this section, we implement subroutine \ope~for tabular MDPs, linear MDPs and  RL problems of low eluder dimension, and derive the respective sample complexity guarantees of \oppo.

\subsection{Tabular MDPs}\label{subsec:tabular}

The implementation of tabular \ope~(Subroutine \ref{alg:tabpe}) is rather standard: first estimate the transition and reward from dataset $\cD$, then plug the estimated transition-reward  into the Bellman equation under policy $\pi$  to compute the Q-value estimate backwards. 
And to guarantee optimism, we additionally add standard counter-based UCB bonus to compensate the error in model estimation. Formally, we have the following guarantee for tabular \ope.

\begin{subroutine}[t]
\caption{Tabular \ope$(\pi, \cD)$}
\label{alg:tabpe}
\begin{algorithmic}[1]
\STATE \textbf{required parameter}: $\alpha$
\STATE split $\cD$ evenly into $H$ disjoint subsets $\cD_1,\ldots,\cD_H$
\STATE compute empirical estimate $(\hat\P_h,\hat R_h)$ of the transition and reward at step $h$ by using $\cD_h$
\STATE  set $\oV_{H+1}(s)\setto 0$ for all $s\in\cS$
\FOR{$h=H:1$}
\STATE for all $(s,a)\in\cS\times\cA$, compute \\
$~\quad J_h(s,a)\setto\sum_{(s_h,a_h)\in\cD_h}\mathbf{1}((s_h,a_h)=(s,a))$\\
and define\vspace{-1mm}
\begin{equation*}\vspace{-2mm}
\begin{cases}
    \oQ_h(s,a)= \min\big\{H-h+1,~\E_{s'\sim\hat\P_h (s,a)}[ \oV_{h+1}(s')]+\hat R_h(s,a)+ b_h(s,a)\big\}  \\
    \oV_h(s)= \E_{a\sim \pi_h(\cdot\mid s)}\left[ \oQ_h(s,a)\right]
    \end{cases}
\end{equation*}
  with bonus function\\ \quad\quad $b_h(s,a)=\alpha({J_h(s,a)+1})^{-1/2}$
  \label{line:tabular-bonus}
\ENDFOR
\STATE \textbf{output}  $\{\oQ_h\}_{h=1}^H$
\end{algorithmic}
\end{subroutine}


\allowdisplaybreaks
\begin{proposition}[tabular MDPs]\label{prop:tabular}
Suppose we choose $\alpha =\Theta\left(H\sqrt{\log(KH|\cS||\cA|)}\right)$ in Subroutine \ref{alg:tabpe}, then Condition \ref{cond:ope} holds with $L=SAH^3$.
\end{proposition}
By combining Proposition \ref{prop:tabular} with 
Theorem \ref{thm:main}, we prove that \oppo~with Subroutine \ref{alg:tabpe} learns an $\epsilon$-optimal policy within $(KN/m)=\tilde{\cO}(H^6|\cS||\cA|/\epsilon^3)$ episodes  for any tabular MDP. 
This rate is strictly worse than the best existing result $\tilde{\cO}(H^3|\cS||\cA|/\epsilon^2)$ in tabular policy optimization \citep{wu2022nearly}. 
Nonetheless, the proof techniques in \citep{wu2022nearly} are specially tailored to tabular MDPs and it is highly unclear how to generalize them to linear MDPs due to certain technical difficulty that arises from  covering a prohibitively large policy space. 
In comparison, our policy optimization framework easily extends to linear MDPs and provides improved sample complexity over the best existing one, as is shown in the following section.

\subsection{Linear MDPs}\label{subsec:linear}

\begin{subroutine}[t]
\caption{Linear $\ope(\pi, \cD)$}
\label{alg:lspe}
\begin{algorithmic}[1]
\STATE \textbf{required parameter}: $\alpha$, $\lambda$
\STATE split $\cD$ evenly into $H$ disjoint subsets $\cD_1,\ldots,\cD_H$
\label{line:split}
\STATE  set $\oV_{H+1}(s)\setto 0$ for all $s\in\cS$
\FOR{$h=H:1$}
\STATE perform ridge regression according to Equation \eqref{eq:ridge}
\label{line:ridge}
\STATE compute covariance matrix\\ ~\quad$\Sigma_h = \sum_{(s_h,a_h)\in\cD_h} \phi_h(s_h,a_h) \phi_h(s_h,a_h)\trans$\\ and define
\begin{equation*}
\begin{cases}
    \oQ_h(s,a)= {\rm Truncate}_{[0,H]}\big(  \langle \hat \theta_h, \phi_h(s,a)\rangle + b_h(s,a)\big) \\
    \oV_h(s)= \E_{a\sim \pi_h(\cdot\mid s)}\left[ \oQ_h(s,a)\right]\vspace{-2mm}
    \end{cases}
\end{equation*}
  with bonus function\\ $~\qquad b_h(s,a):=\alpha\times \|\phi_h(s,a)\|_{(\Sigma_h +\lambda \I_{d\times d})^{-1} }$\label{line:bonus}
\ENDFOR
\STATE \textbf{output}  $\{\oQ_h\}_{h=1}^H$
\end{algorithmic}
\end{subroutine}

We provide the instantiation of \ope~for linear MDPs in Subroutine \ref{alg:lspe}.  At a high level, linear \ope~computes an  upper bound $\overline{Q}$ for $Q^{\pi}$ by using the Bellman equation under policy $\pi$ backwards from step $H$ to step $1$ while adding a bonus to compensate the uncertainty of parameter estimation.
Specifically, the step of ridge regression (Line \ref{line:ridge}) utilizes the linear completeness property of linear MDPs in computing $\overline{Q}_h$ from $\overline{V}_{h+1}$, which states that for any function $\oV_{h+1}:\cS\rightarrow\R$, there exists $\otheta_h\in\R^d$ so that
$\langle \phi_h(s,a),\otheta_h\rangle = R_h(s,a)+\E_{s'\sim\P_h(\cdot\mid s,a)}[\oV_{h+1}(s')]$ for all $(s,a)\in\cS\times\cA$.
And in Line \ref{line:bonus}, we add an elliptical bonus $b_h(s,a)$ to compensate the error between our estimate $\hat\theta_h$ and the groundtruth $\otheta_h$ so that we can  guarantee $\oQ_h(s,a)\ge (\cT_h^\pi \oQ_{h+1})(s,a)$ for all $(s,a,h)$ with high probability .
\begin{equation}
\label{eq:ridge}
\begin{aligned}
\hat \theta_h =  \argmin_{\theta}
\hspace{-6mm}\sum_{(s_h,a_h,r_h,s_{h+1})\in\cD_h} \bigg(\phi_h(s_h,a_h)\trans\theta -  r_h - \oV_{h+1}(s_{h+1}) \bigg)^2 + \lambda \|\theta\|_2^2.
\end{aligned}
\end{equation}

\begin{proposition}[linear MDPs]\label{prop:linear}
Suppose we choose 
$\lambda =1$ and 
$\alpha=\Theta\left(H\sqrt{d\log(KN)}\right)$ in Subroutine \ref{alg:lspe}, then Condition \ref{cond:ope} holds with $L=d^2H^3$.     
\end{proposition}
By combining Proposition \ref{prop:linear} with Theorem \ref{thm:main}, we obtain that $(KN/m)=\tilde{\cO}(d^2H^6/\epsilon^3)$ episodes are sufficient for \oppo~to learn an $\epsilon$-optimal policy, improving upon the state-of-the-art policy optimization results \citep{zanette2021cautiously} by a factor of $d$. 
Notably, this is also the first computationally efficient algorithm to achieve optimal quadratic dimension dependence for learning linear MDPs. 
The key factor behind this improvement is \oppo's periodic collection of fresh on-policy data, which eliminates the undesired correlation and avoids the union bound over certain nonlinear function class that are commonly observed in previous works. 
 Consequently, \oppo~uses a bonus function that is $\sqrt{d}$ times smaller than in previous works  \citep[e.g.,][]{jin2020provably}.
 For further details on how we achieve this improvement, we refer interested readers to Appendix \ref{sec:proof-sketch}, specifically Lemma \ref{lem:opt}.

\subsection{General function approximation}
\label{subsec:eluder}

\begin{subroutine}[t]
\caption{General \ope$(\pi, \cD)$}
\label{alg:eluder}
\begin{algorithmic}[1]
\STATE \textbf{required parameter}: $\beta$
\STATE split $\cD$ evenly into $H$ disjoint subsets $\cD_1,\ldots,\cD_H$
\STATE  set $\oV_{H+1}(s)\setto 0$ for all $s\in\cS$
\FOR{$h=H:1$}
\FOR{$(s,a)\in\cS\times\cA$}
\STATE construct confidence set
\begin{equation*}
    \cB_h =\{f_h\in\cF_h:~\cL_h(\cD_h,f_h) \le  \min_{g_h\in\cF_h}\cL_h(\cD_h,g_h)  + \beta\}\vspace{-2mm}
\end{equation*}
with loss function defined as
\begin{equation*}
    \cL_h(\cD_h,\zeta)= \sum_{(s_h,a_h,r_h,s_{h+1})\in\cD_{h}} \left(\zeta(s_h,a_h)-r_h - \oV_{h+1}(s_{h+1})\right)^2 \vspace{-4mm}
\end{equation*}
  \label{line:conf-set}
\STATE compute
\begin{equation*}
    \begin{cases}
        \oQ_h(s,a) = \sup_{f_h\in\cB_h} f_h(s,a)\\
 \oV_h(s)= \E_{a\sim \pi_h(\cdot\mid s)}\left[ \oQ_h(s,a)\right]
    \end{cases}
\end{equation*}  \label{line:sup-q}

\ENDFOR
\ENDFOR
\STATE \textbf{output}  $\{\oQ_h\}_{h=1}^H$
\end{algorithmic}
\end{subroutine}

Now we instantiate \ope~for RL with  
general function approximation. 
In this setting, the learner is provided with a function class $\cF=\cF_1\times\cdots\times\cF_H$ for approximating the Q values of polices, where $\cF_h\subseteq(\cS\times\cA\rightarrow[0,H])$. 

\paragraph{General \ope.} The pseudocode is provided in Subroutine \ref{alg:eluder}. At each step $h\in[H]$, general \ope~first constructs a confidence set $\cB_h$ which contains candidate estimates for $\cT_h^{\pi}\oQ_{h+1}$ (Line \ref{line:conf-set}). Specifically,  $\cB_h$ consists of all the value candidates $f_h\in\cF_h$ whose square temporal difference (TD) error on dataset $\cD_h$ is no larger than the smallest one by an additive factor $\beta$. Such construction can be viewed as a relaxation of the classic fitted Q-iteration (FQI) algorithm which only keeps the value candidate with the smallest TD error. In particular, if we pick $\beta=0$, $\cB_h$ collapses to the solution of FQI. Equipped with confidence set $\cB_h$, general \ope~then perform optimistic planning at every $(s,a)\in\cS\times\cA$ by computing $\oQ_h(s,a)$ to be the largest possible value that can be achieved by any candidate in confidence set $\cB_h$ (Line \ref{line:sup-q}).  
We remark that general \ope~shares a similar algorithmic spirit to the GOLF algorithm \citep{jin2021bellman}. The key difference is that GOLF is a purely value-based algorithm and only performs optimistic planning at the initial state while general \ope~performs optimistic planning at every state and is part of a  policy gradient algorithm.

\paragraph{Theoretical guarantee.}
To state the main theorem, we need to first introduce two standard concepts: value closeness and eluder dimension, which have been widely used in previous works that study RL with general function approximation. 
\begin{assumption}[value closeness \citep{wang2020reinforcement}]\label{asp:complete}
For all $h\in[H]$ and $V_{h+1}:\cS\rightarrow[0,H]$, $\cT_h V_{h+1} \in\cF_{h}$ where $[\cT_h V_{h+1}](s,a)=R_h(s,a)+\E[V_{h+1}(s')\mid s'\sim\P_h(\cdot\mid s,a)]$. 
\end{assumption}
Intuitively, Assumption \ref{asp:complete} can be understood as requiring that the application of the Bellman operator $\cT_h$ to any V-value function $V_{h+1}$ results in a function that belongs to the value function class $\cF_h$. This assumption holds in various settings, including tabular MDPs and linear MDPs. Furthermore, it is reasonable to assume that value closeness holds whenever the function class $\cF$ has sufficient expressive power, such as the class of neural networks.

\begin{definition}[eluder dimension \citep{russo2013eluder}]
Let $\cG$  be a function class from $\cX$ to $\R$ and $\epsilon>0$. We define the $\epsilon$-eluder dimension of $\cG$, denoted as  $d_E(\cG,\epsilon)$,  to be the largest $L\in\mathbb{N}^{+}$ such that there exists $x_1,\ldots,x_L\in\cX$ and $g_1,g_1',\ldots,g_L,g_L'\in\cG$ satisfying: for all $l\in[L]$, $\sum_{i<l}(g_l(x_i)-g_l'(x_i))^2\le \epsilon$ but $g_l(x_l)-g_l'(x_l)\ge \epsilon$.
\end{definition}
At a high level, eluder dimension $d_E(\cG,\epsilon)$ measures how many mistakes we have to make in order to identify an unknown function from function class  $\cG$ to accuracy $\epsilon$,  in the worst case. It has been widely used as a sufficient condition for proving sample-efficiency guarantee for optimistic algorithms in RL with general function approximation, e.g., \citet{wang2020reinforcement,jin2021bellman}.

Now we state the theoretical guarantee for general \ope. 
\begin{proposition}[general function approximation]\label{prop:eluder}
Suppose Assumption \ref{asp:complete} holds and we choose 
$\beta=\Theta\left(H^2 \log(|\cF|NK/\delta)\right)$ in Subroutine \ref{alg:eluder}, then Condition \ref{cond:ope} holds with $L=H^3 \log(|\cF|)\max_h d_E(\cF_h,1/N)$.     
\end{proposition}
Plugging Proposition \ref{prop:eluder} back into Theorem \ref{thm:main}, we obtain that $(KN/m)=\tilde{\cO}(d_E \log(|\cF|)H^6/\epsilon^3)$ episodes are sufficient for \oppo~to learn an $\epsilon$-optimal policy, which  to our knowledge is the first polynomial sample complexity guarantee for policy optimization with general function approximation.

%% file: Sections/conc.tex

\section{Conclusions}
We proposed a model-free, policy optimization
algorithm---\oppo~for online RL and analyzed its behavior in the episodic setting.
In terms of algorithmic design, it is not only considerably simpler but also more closely resembles  the empirical policy optimization algorithms (e.g., PPO, TRPO) than the previous theoretical algorithms. 
In terms of sample efficiency, for $d$-dimensional linear MDPs, it 
improves over state-of-the-art policy optimization algorithm by a factor of $d$, and is the first computationally efficient algorithm to achieve the optimal dimension dependence. To our best knowledge, \oppo~is also the first sample-efficient policy optimization algorithm under general function approximation. 

For future research, we believe that it is an important direction to investigate the optimal complexity for using policy optimization algorithms to solve linear MDPs. Despite the optimal dependence on dimension $d$, the current sample complexity of \oppo~is worse than the optimal rate by a factor of $1/\epsilon$. It remains unclear to us how to shave off this $1/\epsilon$ factor to achieve the optimal rate.

\section*{Acknowledgement}

Qinghua Liu would like to thank Yu Bai for valuable discussions.
This work was partially supported by National Science Foundation Grant NSF-IIS-2107304.

%% file: Sections/proofs.tex
\section{Proof of Theorem \ref{thm:main}}
\label{sec:proof-sketch}

Recall in Algorithm \ref{alg:oppo}, 
$\piout$ is sampled uniformly at random from $\{\pi^k\}_{k\in[K]}$, so we have 
\begin{align*}
    \lefteqn{\E\left[V^\star_1(s_1) - V_1^{\piout}(s_1)\right]} \\
    & = \frac{1}{K}\sum_{k=1}^K \left[V_1^\star(s_1)-V_1^{\pi^k}(s_1) \right] \\
    & =   
    \underbrace{\frac{1}{K}\sum_{k=1}^K \left[V_1^\star(s_1)-\oV_1^{k}(s_1) \right]}_{\rm \textbf{Term (I)}} +   \underbrace{\frac{1}{K}\sum_{k=1}^K \left[\oV_1^{k}(s_1)-V_1^{\pi^k}(s_1) \right]}_{\rm\textbf{ Term (II)}}.
\end{align*}

To control Term (I), we  import the following generalized  policy difference lemma from \citep{shani2020optimistic,cai2020provably}.
\begin{lemma}[generalized policy-difference lemma]\label{lem:pi-diff}
 For any policy $\pi$ and $k\in[K]$, 
 \begin{equation*}
 \begin{aligned}
V_1^\pi&(s_1) - \oV_1^k (s_1) \\ 
     =  & \sum_{h=1}^H \E_{s_h\sim\pi} \left[\langle \pi_h(\cdot\mid s_h)- \pi^k_h(\cdot\mid s_h),  \oQ^k_h(s_h,\cdot)\rangle  \right] \\
     &- 
     \sum_{h=1}^H \E_{(s_h,a_h)\sim\pi} \Biggl[\oQ_h^k(s_h,a_h) - (\cT_h^{\pi^k} \oQ_{h+1}^k)(s_h,a_h)  \Biggr].
     \end{aligned}
 \end{equation*}
\end{lemma}
By invoking Lemma \ref{lem:pi-diff} with $\pi=\pi^\star$, we have 
\begin{equation}\label{eq:1}
\begin{aligned}
    &{\rm Term (I)} 
    = \frac{1}{K}\sum_{k=1}^K\left[
      V_1^\star(s_1) - \oV_1^k (s_1)  \right]\\
     &\le  H\max_{h,s} \left[\frac{1}{K}\sum_{k=1}^K\left\langle \pi_h^\star(\cdot\mid s)- \pi^k_h(\cdot\mid s),  \oQ^k_h(s,\cdot)\right\rangle  \right] + H\max_{h,s,a}
     \Biggl[ (\cT_h^{\pi^k} \oQ_{h+1}^k -\oQ_h^k)(s,a)  \Biggr] \\
     & \le  \cO\left( \frac{H\log |\cA|}{\eta K} + \eta  H^3\right) +H\max_{h,s,a}
     \Biggl[ (\cT_h^{\pi^k} \oQ_{h+1}^k -\oQ_h^k)(s,a)  \Biggr],
     \end{aligned}
\end{equation}
where the second inequality follows from the standard regret bound of online mirror ascent. 
By Condition \ref{cond:opt} , the second term on the RHS of Equation \eqref{eq:1} is upper bounded by zero. So we have 
$$
{\rm Term (I)}  \le  \cO\left( \frac{H\log |\cA|}{\eta K} + \eta  H^3\right).
$$
As for Term (II), we can directly upper bound it by Condition \ref{cond:consist} which gives:
$$
{\rm Term (II)}  \le  
H\sqrt{(L/N)\times\log(NKL/\delta)}.
$$
We complete the proof by plugging in the choice of $K$, $N$, $\eta$.

\section{Proofs for Examples}

In this section, we prove the three instantiations of \ope~in Section \ref{sec:examples}  satisfy  Condition \ref{cond:ope} with mild complexity measure $L$.
The proofs of the lemmas used in this section can be found in Appendix \ref{app:lem}.

\subsection{Proof of Proposition \ref{prop:linear} (linear \ope)}
We start with the following lemma showing that $\oQ_h^k$ is an optimistic estimate of $\cT_h^{\pi^k} \oQ_{h+1}^k$ and the degree of over-optimism can be bounded by the bonus function. 

\begin{lemma}[optimism of value estimates]\label{lem:opt} Under the same choice of $\lambda$ and $\alpha$ as in Proposition \ref{prop:linear}, with probability at least $1-\delta$: for all $(k,h,s,a)\in[K]\times[H]\times\cS\times\cA$:
\begin{equation*}
    \begin{aligned}
 0\le(\oQ_h^k-\cT_h^{\pi^k} \oQ_{h+1}^k)(s,a)\le 2 b_h^k(s,a),
\end{aligned}
\end{equation*}
where $\{b_h^k\}_{h\in[H]}$ are the bonus functions defined in Line \ref{line:bonus}, Subroutine \ref{alg:lspe} linear $\ope(\pi^k,\cD^k)$.
\end{lemma}
Importantly,  Lemma \ref{lem:opt} holds with a bonus function that  is  smaller by a factor of $\sqrt{d}$ than the typical one used in linear function approximation \citep[e.g.,][]{jin2020provably}, which is key to obtaining the optimal dimension dependence. The main reason that this smaller bonus function suffices is that we recollect fresh samples every $m$ iterations and split the data (Line \ref{line:split} in Subroutine \ref{alg:lspe}) to eliminate the correlation between different steps $h\in[H]$. 

Lemma \ref{lem:opt} immediately implies that Condition \ref{cond:opt} holds. Below we prove Condition \ref{cond:consist}.

By the second inequality in Lemma \ref{lem:opt}, the definition of $\oV^k$ and the Bellman equation under policy $\pi^k$, one can  show 
\begin{equation}\label{eq:2}
  \oV_1^{k}(s_1)-V_1^{\pi^k}(s_1)
  = \sum_{h=1}^H \E_{\pi^k}[(\oQ_h^k-\cT_h^{\pi^k} \oQ_{h+1}^k)(s_h,a_h)]
\le 2\sum_{h=1}^H \E_{\pi^k}[b_h^k(s_h,a_h)].
\end{equation}
Denote by $t_k$ the index of the last iteration of collecting fresh data before the $k\th$ iteration. By exploiting the softmax parameterization structure of $\{\pi^k\}_{k\in[K]}$, one can show that the visitation measure over state-action pairs induced by executing  policy $\pi^k$ is actually very close to the one induced by $\pi^{t_k}$ in the following sense.
\begin{lemma}[policy Lipschitz]\label{lem:pi-close}
Suppose we choose $\eta$ and $m$ such that $\eta m \le 1/H^2$, then for any $k\in\mathbb{N^+}$ and any function $f:\cS\times\cA\rightarrow \R^+$:
$$
\E_{\pi^k}[f(s_h,a_h)] = \Theta\left(
\E_{\pi^{t_k}}[f(s_h,a_h)]\right).
$$
\end{lemma}
By combining Lemma \ref{lem:pi-close} with Equation \eqref{eq:2} and noticing that $b_h^k=b_h^{t_k}$, we have 
$$ \oV_1^{k}(s_1)-V_1^{\pi^k}(s_1) \le  \cO\bigg(\max_{k\in[K]}\sum_{h=1}^H \E_{\pi^{t_k}}[b_h^k(s_h,a_h)]\bigg)=\cO\bigg(\max_{k\in[K]}\sum_{h=1}^H \E_{\pi^{t_k}}[b_h^{t_k}(s_h,a_h)]\bigg).
    $$
Since the bonus function $b_h^{t_k}$ is defined by using the dataset $\cD^{t_k}$ collected from executing policy $\pi^{t_k}$, one could expect its average under $\pi^{t_k}$ should be rather small when the batch size $N$ is large enough, as formalized in the following lemma.

\begin{lemma}[on-policy uncertainty] \label{lem:on-policy-bonus}
Let $\pi$ be an arbitrary policy and $\lambda \ge 1$. 
Suppose we sample $\{(s_h^n,a_h^n)\}_{n=1}^N$ i.i.d. from $\pi$. Denote $\Sigma_h=\sum_{i=1}^{N} \phi_h(s_h^i,a_h^i) \phi_h\trans(s_h^i,a_h^i)$. 
Then with  probability at least $1-\delta$:
\begin{equation*}
\begin{aligned}
    &\E_{(s_h,a_h)\sim\pi}\big[\|\phi_h(s_h,a_h)\|_{(\Sigma_h+ \lambda  \I_{d\times d})^{-1}}\big] = \cO\left(\sqrt{\frac{d\log (N/\delta)}{N} }\right).
\end{aligned}
\end{equation*}
\end{lemma}
As a result, by Lemma \ref{lem:on-policy-bonus} and the definition of $b_h^k$, we have that with  probability at least $1-\delta$: for all $k\in[K]$, 
$$
\sum_{h=1}^H \E_{\pi^{t_k}}[b_h^k(s_h,a_h)] = {\cO}\left( Hd\sqrt{\frac{H\log^2(NKH/\delta)}{N}}\right)
$$
where in applying Lemma \ref{lem:on-policy-bonus}  we use the fact $|\cD_h^k|=N/H$ due to data splitting. 
Putting all pieces together, we get
$$
\oV_1^{k}(s_1)-V_1^{\pi^k}(s_1) \le {\cO}\left( \sqrt{\frac{d^2 H^3\log^2(KNH/\delta)}{N}}\right).
$$
As a result, Condition \ref{cond:consist} holds with $L=d^2H^3$.

\subsection{Proof of Proposition \ref{prop:tabular} (tabular \ope)}

The proof  follows basically the same as that of Proposition \ref{prop:linear}. The only modification needed is to replace Lemma \ref{lem:opt} and \ref{lem:on-policy-bonus} by their tabular counterparts, which we state below. 

\begin{lemma}[optimism of value estimates: tabular]\label{lem:opt-tabular} Under the same choice of $\alpha$ as in Proposition \ref{prop:tabular}, with probability at least $1-\delta$: for all $(k,h,s,a)\in[K]\times[H]\times\cS\times\cA$:
\begin{equation*}
    \begin{aligned}
 0\le(\oQ_h^k-\cT_h^{\pi^k} \oQ_{h+1}^k)(s,a)\le 2 b_h^k(s,a),
\end{aligned}
\end{equation*}
where $\{b_h^k\}_{h\in[H]}$ are the bonus functions defined in Line \ref{line:tabular-bonus}, Subroutine \ref{alg:tabpe} tabular $\ope(\pi^k,\cD^k)$.
\end{lemma}

\begin{lemma}[on-policy uncertainty: tabular] \label{lem:on-policy-bonus-tabular}
Let $\pi$ be an arbitrary policy. 
Suppose we sample $\{(s_h^n,a_h^n)\}_{n=1}^N$ i.i.d. from $\pi$. Denote $J_h(s,a)=\sum_{i=1}^{N} \mathbf{1}((s_h^i,a_h^i)=(s,a))$. 
Then with  probability at least $1-\delta$:
\begin{equation*}
\begin{aligned}
    \E_{(s_h,a_h)\sim\pi}\left[\sqrt{\frac{1}{J_h(s,a)+1}}\right] = \cO\left(\sqrt{\frac{SA\log (N/\delta)}{N} }\right).
\end{aligned}
\end{equation*}
\end{lemma}

\subsection{Proof of Proposition \ref{prop:eluder} (general \ope)}
\label{subsec:elude}

The proof  follows basically the same as that of Proposition \ref{prop:linear}. The only modification needed is to replace Lemma \ref{lem:opt} and \ref{lem:on-policy-bonus} by their general counterparts, which we state below. 

\begin{lemma}[optimism of value estimates: general]\label{lem:opt-eluder} Suppose Assumption \ref{asp:complete} holds. 
There exists an absolute constant $c$ such that  with probability at least $1-\delta$: for all $(k,h,s,a)\in[K]\times[H]\times\cS\times\cA$:
\begin{equation*}
    \begin{aligned}
 0\le(\oQ_h^k-\cT_h^{\pi^k} \oQ_{h+1}^k)(s,a)\le  b_h(s,a,\cD_h^k):=
\end{aligned}
\left(
\begin{aligned}
   & \sup_{f_h,f_h'\in\cF_h}f_h(s,a)-f_h'(s,a) \\
    \text{ s.t. }
 &\sum_{(s_h,a_h)\in\cD_h^k}(f_h(s_h,a_h)-f_h'(s_h,a_h))^2\le c\beta
\end{aligned}\right).
\end{equation*}
\end{lemma}

\begin{lemma}[on-policy uncertainty: general] \label{lem:on-policy-bonus-eluder}
Let $\pi$ be an arbitrary policy. 
Suppose we sample $\cD_h=\{(s_h^n,a_h^n)\}_{n=1}^N$ i.i.d. from $\pi$. 
Then with  probability at least $1-\delta$:
\begin{equation*}
\begin{aligned}
\E_{(s_h,a_h)\sim\pi}\left[b_h(s,a,\cD_h) \right] = \cO\left(H\sqrt{\frac{d_E(1/N,\cF_h)\beta}{N} }+ H\sqrt{\frac{\log(1/\delta)}{N}}\right).
\end{aligned}
\end{equation*}
\end{lemma}

%% file: Sections/app.tex
\section{Proofs of Lemmas}
\label{app:lem}

\subsection{Proof of Lemma \ref{lem:pi-diff}}
Lemma \ref{lem:pi-diff} is an immediate consequence of Lemma 4.2 in \citep{cai2020provably} or Lemma 1 in \citep{shani2020optimistic}.

\subsection{Proof of Lemma \ref{lem:opt}}
Let us consider a fixed pair  $(k,h)\in[K]\times[H]$. Recall $\oQ_h^k$ is defined as 
$$
\oQ_h^k(s,a)= {\rm Truncate}_{[0,H-h+1]}\big( \langle \hat \theta_h^k, \phi_h(s,a)\rangle + b_h^k(s,a)\big), 
$$
where 
\begin{equation*}
\hat \theta_h^k =  \argmin_{\theta}
\hspace{-6mm}\sum_{(s_h,a_h,r_h,s_{h+1})\in\cD_h^k} \bigg(\phi_h(s_h,a_h)\trans\theta  -  r_h - \oV_{h+1}^k(s_{h+1}) \bigg)^2 + \lambda \|\theta\|_2^2,
\end{equation*}
\begin{equation*}
    b_h^k(s,a) = \alpha\times \|\phi_h(s,a)\|_{(\Sigma_h^k +\lambda \I_{d\times d})^{-1} }, \quad \text{and} \quad \Sigma_h^k =\sum_{(s_h,a_h)\in\cD_h^k} \phi(s_h,a_h)(\phi(s_h,a_h))\trans.
\end{equation*}
By the standard linear completeness property of linear MDPs \cite{jin2020provably}, there exists $\theta_h^k$ such that 
$\|\theta_h^k\|_2\le H\sqrt{d}$ and 
$$
\langle \phi_h(s,a), \theta_h^k\rangle = R_h(s,a) + \E_{s'\sim\P(\cdot \mid s,a)}\left[ \oV_{h+1}^k(s')\right] \quad \text{for all } (s,a)\in\cS\times\cA.
$$
Therefore, to prove Lemma \ref{lem:opt}, it suffices to show that  for all $(s,a)$
$$
\left|\langle \phi_h(s,a),\theta_h^k -
\hat\theta_h^k\rangle\right|\le b_h^k(s,a).
$$
To condense notations, denote by $\{(s_h^i,a_h^i,r_h^i,s_{h+1}^i)\}_{i=1}^M$ the tuples in $\cD_h^k$. 
By the definition of ridge regression,
\begin{align*}
   \MoveEqLeft \left|\langle \phi_h(s,a),\hat\theta_h^k-\theta_h^k \rangle\right| \\
=&\bigg|\bigg\langle \phi_h(s,a)~,~(\Sigma_h^k+\lambda \I_{d\times d})^{-1}\times\\
&\left(\sum_{i=1}^M\phi_h(s_h^i,a_h^i) \left(  r_h^i + \oV_{h+1}^k(s_{h+1}^i)-\phi_h(s_h^i,a_h^i)\trans\theta^k_h\right)-\lambda \theta_h^k\right)\bigg\rangle\bigg|\\
\le & \|\phi_h(s,a)\|_{(\Sigma_h^k +\lambda \I_{d\times d})^{-1} } \times \\ &\left(\left\|\sum_{i=1}^M\phi_h(s_h^i,a_h^i) \left(  r_h^i + \oV_{h+1}^k(s_{h+1}^i)-\phi_h(s_h^i,a_h^i)\trans\theta^k_h\right)\right\|_{(\Sigma_h^k +\lambda \I_{d\times d})^{-1} }+H\sqrt{d} \right),
\end{align*}
where the inequality follows from 
$ \| \lambda \theta_h^k\|_{(\Sigma_h^k +\lambda \I_{d\times d})^{-1} }\le \sqrt{\lambda}\|\theta_h^k\|_2 \le H\sqrt{d}.
$
It remains to bound the first term in the bracket.
Now here comes the key observation that helps shave off the $\sqrt{d}$ factor from the bonus.
\begin{observation}\label{obs:ind}
$\cD_h^k$ and $\oV_{h+1}^k$ are independent conditioning on $\pi^{t_k}$ where $t_k$ denotes the index of the last iteration of collecting fresh data before the $k\th$ iteration.
\end{observation}
To see why, recall that after obtaining $\cD^{t_k}$ Algorithm \ref{alg:lspe} splits it evenly into $H$ disjoint subsets $\{\cD_h^{t_k}\}_{h=1}^H$ and then only uses
$\{\cD_{h'}^{t_k}\}_{h'=h+1}^H$ for evaluating and improving $\{\pi_{h'}^l\}_{h'=h+1}^H$ for $l\in[t_k,t_k+m-1]$. As a result,  $\{\oV_{h+1}^l\}_{l=t_k}^{t_k+m-1}$ (including $\oV_{h+1}^k$) is independent of $\cD_h^{t_k}=\cD_h^k$ conditioning on $\pi^{t_k}$. 
By the concentration of self-normalized processes \citep{abbasi2011improved}, we conclude that with probability at least $1-\delta$: for all $(k,h)\in[K]\times[H]$
$$
\left\|\sum_{i=1}^M\phi_h(s_h^i,a_h^i) \left(  r_h^i + \oV_{h+1}^k(s_{h+1}^i)-\phi_h(s_h^i,a_h^i)\trans\theta^k_h\right)\right\|_{(\Sigma_h^k +\lambda \I_{d\times d})^{-1} }
\le \cO\left(H\sqrt{d\log(MKH/\delta)}\right).
$$

\subsection{Proof of Lemma \ref{lem:pi-close}}

We first introduce the following two auxiliary lemmas about the lipschitzness continuity of softmax paraterization.

\begin{lemma}\label{lem:policy-stability}
There exists an absolute constant $c>0$ such that for any policy $\pi,\hat\pi$ satisfying $\hat\pi_h(a\mid s) \propto \pi_h(a\mid s)\times \exp(L_h(s,a))$ where $\{L_h(s,a)\}_{h\in[H]}$ is set of  functions from $\cS\times\cA$ to $[-1/H,1/H]$, we have that for any $\tau_H:=(s_1,a_1,\ldots,s_H,a_H)\in(\cS\times\cA)^H$:
$$
\P^{\hat\pi}(\tau_H) 
\le c\times \P^\pi(\tau_H).
$$
\end{lemma}
\begin{proof}[Proof of Lemma \ref{lem:policy-stability}]
By using the relation between $\pi$ and $\hat\pi$ and the normalization property of policy $\pi$, we have that for any $(h,s,a)$:
\begin{align*}
\hat\pi_h(a\mid s)
& = \frac{\pi_h(a\mid s) \times \exp(L_h(s,a))}{\sum_{a'} \pi_h(a'\mid s) \times \exp(L_h(s,a'))} \\
& \le 
\frac{\pi_h(a\mid s) \times \exp(1/H)}{\sum_{a'} \pi_h(a'\mid s) \exp(-1/H) } \\
& = \pi_h(a\mid s) \times \exp(2/H)
\le \pi_h(a\mid s) \times \left(1+\frac{\cO(1)}{H}\right).
\end{align*}
Therefore, for any 
$\tau_H:=(s_1,a_1,\ldots,s_H,a_H)$
\begin{align*}
\P^{\hat\pi}(\tau_H) = &  \left(\prod_{h=1}^{H-1} \P_h(s_{h+1}\mid s_h,a_h)\right)
\times 
\left(\prod_{h=1}^{H} \hat\pi_h(a_h\mid s_h)\right)\\
\le  & \left(\prod_{h=1}^{H-1} \P_h(s_{h+1}\mid s_h,a_h)\right)
\times 
\left(\prod_{h=1}^{H}
\pi_h(a_h\mid s_h)\right)\times \left(1+\frac{\cO(1)}{H}\right)^H \\
= &\cO(1) \times 
\left(\prod_{h=1}^{H-1} \P_h(s_{h+1}\mid s_h,a_h)\right)
\times 
\left(\prod_{h=1}^{H}
\pi_h(a_h\mid s_h)\right)
= \cO(1) \times 
\P^{\pi}(\tau_H).
\end{align*}
\end{proof}

\begin{lemma}\label{lem:transfer}
Let $\mu,\nu$ be two probability densities defined over $\cX$ such that 
$\|\mu/\nu\|_\infty \le \alpha$, then we have that  for any function  $f:\cX\rightarrow \R^+$, $\E_\mu[f(x)] \le \alpha \E_\nu[f(x)]$.
\end{lemma}
\begin{proof}[Proof of Lemma \ref{lem:transfer}]
By definition, $\E_\mu[f(x)] = \int_x f(x) \mu(x)dx \le \int_x f(x) (\alpha\nu(x))dx = \alpha \E_\nu[f(x)].$
\end{proof}

By the update rule of NPG, we have 
$$
\pi^{k}_h(\cdot \mid s) \propto \pi^{t_k}_h(\cdot \mid s)\times \exp\left( \eta\sum_{i=t_k}^{k-1} \oQ_h^i(s,\cdot) \right).
$$
Since we choose $\eta$ and $m$ such that $\eta m \le 1/H^2$, 
$$
\left|\eta\sum_{i=t_k}^{k-1} \oQ_h^i(s,\cdot)\right|\le \eta (k-t_k)H\le \eta m H \le 1/H.
$$
Therefore, by invoking Lemma \ref{lem:policy-stability} with $\hat\pi=\pi^k$ and $\pi=\pi^{t_k}$, we have that for any $\tau_H\in(\cS\times\cA)^H$:
$$
\P^{\pi^k}(\tau_H) 
\le c\times \P^{\pi^{t_k}}(\tau_H).
$$
By further invoking Lemma \ref{lem:transfer} with $\cX=(\cS\times\cA)^H$, $\mu=\P^{\pi^k}$ and $\nu=\P^{\pi^{t_k}}$, we obtain that for any function $f:\cS\times\cA\rightarrow \R^+$:
$$
\E_{\pi^k}[f(s_h,a_h)] = \cO\left(
\E_{\pi^{t_k}}[f(s_h,a_h)]\right).
$$
Similarly, we can show 
$\E_{\pi^{t_k}}[f(s_h,a_h)] = \cO\left(\E_{\pi^k}[f(s_h,a_h)]\right)$, which completes the proof of Lemma \ref{lem:pi-close}.

\subsection{Proof of Lemma \ref{lem:on-policy-bonus}}

To simplify notations, denote  $\phi^i_h :=\phi_h(s_h^i,a_h^i)$.
For the technical purpose of applying martingale concentration, we additionally define the intermediate empirical covariance matrices: $\Sigma_h^n=\sum_{i=1}^{n-1} \phi_h^i (\phi_h^i)\trans$, for $n\in[N]$. 
We have that with probability at least $1-\delta$,
\begin{equation}
    \begin{aligned}
     \MoveEqLeft \E_{(s_h,a_h)\sim\pi}\big[\|\phi_h(s_h,a_h)\|_{(\Sigma_h+ \lambda  \I_{d\times d})^{-1}}\big] \\ &\le \frac{1}{N}\sum_{n=1}^N  \E_{(s_h,a_h)\sim\pi}\big[\|\phi_h(s_h,a_h)\|_{(\Sigma_h^n+ \lambda  \I_{d\times d})^{-1}}\big] \\
     & \le \frac{1}{N}\sum_{n=1}^N  \|\phi_h^n\|_{(\Sigma_h^n+ \lambda  \I_{d\times d})^{-1}} + \cO\left( \sqrt{\frac{\log(1/\delta)}{N}}\right)\\
     & \le \cO\left(\sqrt{\frac{d\log N}{N}}\right) + \cO\left( \sqrt{\frac{\log(1/\delta)}{N}}\right), 
    \end{aligned}
\end{equation}
where the first inequality uses $\Sigma_h^n \preceq \Sigma_h$,  the second one uses Azuma-Hoeffding inequality with $\lambda =1$, and the last one uses Cauchy–Schwarz inequality and the the standard elliptical potential argument.

\subsection{Proof of Lemma \ref{lem:opt-tabular}} Let us consider a fixed pair  $(k,h)\in[K]\times[H]$. Recall $\oQ_h^k$ is defined as 
\begin{equation*}
    \oQ_h^k(s,a)= \min\big\{H-h+1,~
    \E_{s'\sim\hat\P_h^k (s,a)}[ \oV_{h+1}^k(s')]+\hat R_h^k(s,a)+ b_h^k(s,a)\big\}
\end{equation*}
 where  $\hat\P_h^k$ and $\hat\R_h^k$ are the empirical estimates of transition and reward at step $h$ from $\cD_h^k$, and 
 $b_h^k(s,a)=\alpha({J_h^k(s,a)+1})^{-1/2}$ with $J_h^k(s,a)=\sum_{(s_h,a_h)\in\cD_h^k}\mathbf{1}((s_h,a_h)=(s,a))$. Therefore, to prove Lemma \ref{lem:opt-tabular}, it suffices to show that  for all $(s,a)$
$$
\left|\left(\E_{s'\sim\hat\P_h^k (s,a)}[ \oV_{h+1}^k(s')] - 
\E_{s'\sim\P_h (s,a)}[ \oV_{h+1}^k(s')]\right)+\left(\hat R_h^k(s,a) - R_h(s,a)\right)
\right|\le b_h^k(s,a).
$$
By observation \ref{obs:ind}, we know    $\{\oV_{h+1}^l\}_{l=t_k}^{t_k+m-1}$ is independent of $\cD_h^{t_k}=\cD_h^k$ conditioning on $\pi^{t_k}$, which implies $\oV_{h+1}^k$ is independent of $\hat\P_h^k$. Therefore, by Azuma-Hoeffding inequality and standard union bound, we conclude that with probability at least $1-\delta$: for all $(k,h,s,a)\in[K]\times[H]\times\cS\times\cA$, 
\begin{align*}
\left|\left(\E_{s'\sim\hat\P_h^k (s,a)}[ \oV_{h+1}^k(s')] - 
\E_{s'\sim\P_h (s,a)}[ \oV_{h+1}^k(s')]\right)+\left(\hat R_h^k(s,a) - R_h(s,a)\right)
\right|\\
\le \cO\left(H\sqrt{\frac{\log(KNHSA/\delta)}{J_h^k(s,a)+1}}\right).    
\end{align*}

\subsection{Proof of Lemma \ref{lem:on-policy-bonus-tabular}}For the technical purpose of performing martingale concentration, we introduce the notion of  intermediate counters: $J_h^n(s,a)=\sum_{i=1}^{n-1}  \mathbf{1}((s_h^i,a_h^i)=(s,a))$, for $n\in[N]$. 
We have that with probability at least $1-\delta$,
\begin{equation}
    \begin{aligned}
     \MoveEqLeft   \E_{(s_h,a_h)\sim\pi}\left[\sqrt{\frac{1}{J_h(s,a)+1}}\right]  \\ &\le \frac{1}{N}\sum_{n=1}^N    \E_{(s_h,a_h)\sim\pi}\left[\sqrt{\frac{1}{J_h^n(s,a)+1}}\right]  \\
     & \le \frac{1}{N}\sum_{n=1}^N  \sqrt{\frac{1}{J_h^n(s_h^n,a_h^n)+1}}+ \cO\left( \sqrt{\frac{\log(1/\delta)}{N}}\right)\\
     & \le \cO\left(\sqrt{\frac{SA\log N}{N}}\right) + \cO\left( \sqrt{\frac{\log(1/\delta)}{N}}\right), 
    \end{aligned}
\end{equation}
where the first inequality uses $J_h^n(s,a) \le J_h(s,a)$ for all $(s,a)\in\cS\times\cA$,  the second one uses Azuma-Hoeffding inequality, and the last one follows the standard pigeon-hole argument.

\subsection{Proof of Lemma \ref{lem:opt-eluder}}
By observation \ref{obs:ind}, we know  
 $\cD_h^k$ is  independent of $\oV_{h+1}^k$  conditioning on $\pi^{t_k}$ where $t_k$ is the index of the last iteration when \oppo~collects fresh data before iteration $k+1$. With this independence relation in mind, we can easily prove that confidence set $\cB_h^k$ satisfy the following properties by standard concentration inequality and union bounds. 
\begin{lemma}\label{lem:concen}
Suppose Assumption \ref{asp:complete} holds. There exists an absolute constant $c$ such that  with probability at least $1-\delta$, for all $k\in[K]$ and $h\in[H]$,
\begin{itemize}
    \item $\cT_h^{\pi^k} \oQ_{h+1}^k\in\cB_h^k$,
    \item for any $f_h\in\cB_h^k$, $\sum_{(s_h,a_h)\in\cD_h^k}(f_h(s_h,a_h)-\cT_h^{\pi^k} \oQ_{h+1}^k(s_h,a_h))^2 \le c\beta$.
\end{itemize}
\end{lemma}
The proof of Lemma \ref{lem:concen} follows  trivially from modifying the proofs of Lemma 39 and 40 in \citet{jin2021bellman}, which we omit here. 

Define 
$$
\bar\cB_h^k = \{f_h\in\cF_h:~\sum_{(s_h,a_h)\in\cD_h^k}(f_h(s_h,a_h)-(\cT_h^{\pi^k} \oQ_{h+1}^k)(s_h,a_h))^2\le c\beta\}.
$$
By using the first relation in Lemma \ref{lem:concen} and the definition of $\oQ_h^k$, we immediately obtain that 
\begin{align*}
(\cT_h^{\pi^k} \oQ_{h+1}^k)(s,a) \le 
\sup_{f_h\in\cB_h^k} f_h(s,a) = \oQ_h^k(s,a) 
\end{align*}
and 
\begin{align*}
\oQ_h^k(s,a)  -(\cT_h^{\pi^k} \oQ_{h+1}^k)(s,a) 
\le \sup_{f_h,f'_h\in\cB^k_h}(f_h(s,a)-f_h'(s,a)) \le \sup_{f_h,f'_h\in\bar\cB^k_h}(f_h(s,a)-f_h'(s,a)),
\end{align*}
where the last inequality uses $\bar\cB_h^k\subset \cB_h^k$ by Lemma \ref{lem:concen}.

\subsection{Proof of Lemma \ref{lem:on-policy-bonus-eluder}}
Recall  we define 
\begin{equation*}  
b_h(s,a,\cD_h):=
\left(
\begin{aligned}
   & \sup_{f_h,f_h'\in\cF_h}f_h(s,a)-f_h'(s,a) \\
    \text{ s.t. }
 &\sum_{(s_h,a_h)\in\cD_h}(f_h(s_h,a_h)-f_h'(s_h,a_h))^2\le c\beta
\end{aligned}\right),
\end{equation*}
which implies that $b_h(s,a,\cD_h)\le b_h(s,a,\underline{\cD}_h)$ for any  $\underline{\cD}_h\subseteq \cD_h$. 
Let $\cD_h^{(n)}=\{(s_h^i,a_h^i)\}_{i=1}^n$. We have that with probability at least $1-\delta$,
\begin{align*}
\E_{(s_h,a_h)\sim\pi}\left[b_h(s,a,\cD_h) \right] 
    &\le \frac{1}{N}\sum_{n=1}^N \E_{(s_h,a_h)\sim\pi}\left[b_h(s,a,\cD_h^{(n-1)}) \right]  \\
    &\le  \frac{1}{N}\sum_{n=1}^N b_h(s_h^n,a_h^n,\cD_h^{(n-1)})  +  \cO\left( H\sqrt{\frac{\log(1/\delta)}{N}}\right)\\
    &\le \cO\left(H\sqrt{\frac{d_E(1/N,\cF_h)\beta}{N} }+ H\sqrt{\frac{\log(1/\delta)}{N}}\right),
\end{align*}
where the second inequality uses Azuma-Hoeffding inequality  and the last one uses the standard regret guarantee for eluder dimension (e.g., Lemma 2 in \citet{russo2013eluder}).